\documentclass[sigconf]{acmart}
\usepackage{graphicx}
\usepackage{natbib}
\usepackage{xcolor}
\usepackage{subfigure}
\usepackage{url}
\usepackage{algpseudocode}
\usepackage{algorithm}
\usepackage{colortbl}

\graphicspath{ {images/} }
\newcommand{\algrule}[1][.2pt]{\par\vskip.5\baselineskip\hrule height #1\par\vskip.5\baselineskip}

\renewcommand\footnotetextcopyrightpermission[1]{}
\setcopyright{none}
\title{User Friendly  Automatic Construction of Background Knowledge: \\Mode Construction from ER Diagrams}
\author{Alexander L. Hayes}
\affiliation{University of Texas at Dallas\\Alexander.Hayes@utdallas.edu}
\author{Mayukh Das}
\affiliation{University of Texas at Dallas\\Mayukh.Das1@utdallas.edu}
\author{Phillip Odom}
\affiliation{Indiana University, Bloomington\\phodom@indiana.edu}
\author{Sriraam Natarajan}
\affiliation{University of Texas at Dallas\\ Indiana University. Bloomington\\Sriraam.Natarajan@utdallas.edu}
\date{}

\begin{document}
\newcommand{\hlightred}[1]{\colorbox{red!25}{$\displaystyle#1$}}
\newcommand{\hlightblue}[1]{\colorbox{blue!25}{$\displaystyle#1$}}
\newcommand{\hlightgreen}[1]{\colorbox{green!25}{$\displaystyle#1$}}
\newcommand{\hlightpurp}[1]{\colorbox{purple!25}{$\displaystyle#1$}}
\begin{abstract}

One of the key advantages of Inductive Logic Programming systems is the ability of the domain experts to provide background knowledge as modes that allow for efficient search through the space of hypotheses. However, there is an inherent assumption that this expert should also be an ILP expert to provide effective \textit{modes}. We relax this assumption by designing a graphical user interface that allows the domain expert to interact with the system using Entity Relationship diagrams. These interactions are used to construct modes for the learning system. We evaluate our algorithm on a probabilistic logic learning system where we demonstrate that the user is able to construct effective background knowledge on par with the expert-encoded knowledge on five data sets.

\end{abstract}

\maketitle

\section{Introduction}
Recently, there has been an increase in the development of algorithms and models that combine the expressiveness of first-order logic with the ability of probability theory to model uncertainty. Collectively called {Probabilistic Logic Models} (PLMs) or {Statistical Relational Learning} models (SRL)~\cite{srlBook,starAIbook}, these methods have become popular for learning in the presence of multi-relational noisy data. While effective, learning these models remains a computationally intensive task. This is due to the fact that the learner should search for hypotheses at multiple levels of abstraction.

Consequently, methods whose search strategies are inspired from Inductive Logic Programming (ILP) have been introduced to make learning more efficient~\cite{natarajan2015boosted,natarajan10}. These methods have demonstrated arguably some of the best results in several benchmark and real data sets. While effective, the key issue with these methods is that they require the domain expert to also be an expert in ILP---thus providing the right set of directives for learning the target concepts. These additional directives, typically
called {\em modes}, restrict the search space such that the learning of these probabilistic clauses is efficient. Many real users of these systems, especially those who fail to learn good models with these algorithms, may not able to select the correct modes to guide the search.
The consequence is that many of the learning procedures get stuck in a local minimum or get timed out resulting in sub-optimal models.

One way that this problem has been addressed in literature is by employing databases underneath the learner to improve the search speed~\cite{malec2016inductive,zeng2014quickfoil,niuvldb11}. While these systems have certainly improved the search, recent work by Malec et al.~\cite{malec2016inductive} clearly demonstrated the need for modes to achieve effective learning even when using databases. Their work showed an order of magnitude improvement over the standard state-of-the-art PLM learning system. However, their work also required the modes to be specified for attaining this efficiency.

Inspired by their success, we propose a method for specifying modes from a database perspective. Specifically, we propose to employ the use of Entity Relationship (ER) diagrams  as the graphical tools based on which an user could specify modes. The key intuition is that the modes specify how the search is conducted through the space of hypotheses. When viewed from a relational perspective, this can be seen as specifying the parts of the relational graph that are relevant to the target concept. We provide an interface that allows for an user to guide the PLM learners using ER diagrams. Our interface automatically converts the user inputs on ER diagrams to mode definitions that are then later employed to guide the search. Our work is also inspired by the work of Walker et al.~\cite{walker2011integrating} where a UI was designed to provide {\em advice} for
a PLM learner. While their UI was domain-specific, our contribution is a generalized approach to utilize any ER diagram to automatically construct background knowledge for logic-based learners. Our work is also related to the other work of Walker et al.~\cite{walker2010automating} where the mode construction was automated using a layered approach which relied on successively broadening the search space until a relevant model was found.  While their work was effective, due to the layering, scaling their work to large tasks can be inefficient. Ours is a more restricted approach which allows for a domain expert to specify the modes using an ER diagram.

We make the following contributions: (1) We propose an approach to make ILP and PLM systems more usable by domain experts by creating a graphical user interface. (2) We demonstrate how effective background knowledge can be encoded using an ER diagram and provide an algorithm for the  translation from UI input to a mode specification file. (3) We show empirically the effectiveness of our learning approach in standard PLM tasks.

We first provide the necessary background knowledge of ER diagrams and ILP systems. Then we outline the procedure for converting the ER inputs to mode definitions. We finally conclude the paper by demonstrating the effectiveness on standard PLM domains and outlining the directions for future research.
\section{Background}

As our approach uses ER diagrams as a way of constructing background knowledge through modes, we discuss both the diagram as well as how modes are typically used in ILP.

\subsection{The Entity-Relational Model}

\begin{figure}
    \centering
    \includegraphics[width=\columnwidth]{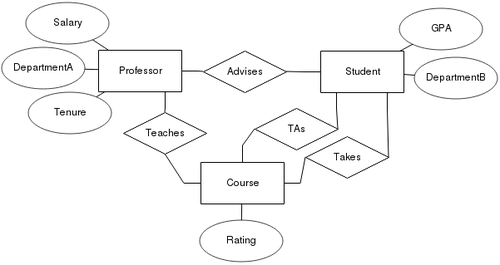}
    \caption{An ER-Diagram illustrating 3 entities, \textit{Professors, Students, and Courses}, their attributes (ovals)  and the relationships (diamonds) among them.}
    \label{fig:egERD}
\end{figure}

The entity-relationship model~\cite{chen1976entity} allows for expressing the structure and semantics of a database at an abstract level as objects and classes of objects (entities and entity classes), attributes of such entities, and relationships that exist between such entity classes. Entities are represented as rectangles, attributes as circles and relationships as diamonds. While relational logic (as used in ILP) is equally expressive, ER models are represented as graphical structures (ER diagrams) making them more intuitive and interpretable. ER modeling is insufficient for expressing operations on the data, but in our problem setting that has no impact. Figure~\ref{fig:egERD} illustrates an ER diagram for an example domain.

ER diagrams are commonly utilized by both database designers and domain experts to conceptualize the structural characteristics of a given domain~\cite{garcia2009database}. A relational schema is an alternate abstract representation of the structure of relational data consisting of structural definitions of relational tables, attributes, foreign key constraints, etc. Conceptually, the knowledge conveyed by a relational schema can be used for abstracting background/modes for an ILP/PLM learning task. But, the limitation lies in the ambiguity it may introduce. For instance, the relation node \textit{``TAs''} (Figure~\ref{fig:egERD}) can be expressed in a relational schema either as a foreign key constraint from one entity to another or as a table with 2 columns having the unique identifiers of the connected entities \textit{Course} and \textit{Student}. The choice typically depends on the design of the system that will use the database. ER diagrams avoid such ambiguity via consistent syntax.

Our approach is motivated from the intuitive connections between constrained logical clause search and SQL (Structured Query Language) query augmentation.
Logical clauses are equivalent to relational queries since, fundamentally, SQL statements are manifestations of entity sets defined via relational calculus. Several ILP/PLM learning frameworks have successfully utilized this concept~\cite{niuvldb11,malec2016inductive}. Similarly, modes for clause learning can be interpreted as constraints on relational query construction and query evaluation.  ``Hints," in relational queries, are special symbolic tools to guide the query evaluation engine to prioritize some database operation over the others to enhance efficiency~\cite{lohman2000relational,diab2009search,bruno2012flexible}. Thus, they are akin to soft directives/constraints (modes) on the search space.

\subsection{Background Knowledge for ILP}

Background knowledge serves two purposes in ILP systems: describing the underlying structure of data and constraining the space of models over which the algorithm  explores. Thus, background knowledge (set via modes) is a key component for getting relational learning algorithms to work effectively. A mode describes a way of instantiating predicates in a clause that defines a hypothesis. A mode for predicate $pred$ with $n$ arguments is defined as $pred(type1,type2,...,typen)$. Each type describes the domain of objects which can appear as that argument, as well as whether it can be instantiated with an input variable (+), an output variable (-), or a constant (\#)~\cite{aleph}. Input variables must be previously defined in the model. Output variables are free variables that have not been defined.

ILP learners search through the space of models in different ways. Aleph~\cite{aleph} generates clauses bottom-up by constructing the most specific explanation of examples and then generalizing while TILDE~\cite{tilde} constructs clauses top-down. Our mode construction approach is capable of generating background knowledge for a variety of different ILP systems. To validate our approach, we make use of a state-of-the-art ILP system called Relational Functional Gradient Boosting (RFGB)~\cite{boostingMLJ12} that learns a set of boosted relational regression trees in a top-down manner. Relational regression trees contain relational logic in the inner nodes and regression values in the leaves. Each iteration of RFGB learns a tree ($\psi_k$) that pushes the model in the direction of the current error. The error of the current model ($\Delta_{k-1}$) is computed over each training example: $\Delta_{k-1}(x_i) = I(y_i=1) - P(y_i=1| Pa(x_i))$ where $I$ is an indicator function for whether $x_i$ is a positive example and $P$ represents the current predicted probability. The final model is a sum over all of trees ($\psi_M=\psi_0+\psi_1+...+\psi_m$). For more details we refer to Natarajan et al.~\cite{boostingMLJ12}.

\begin{algorithm}[t!]
\caption{Guided Mode Construction (\textsc{GMC})}
\label{algo:gmc}
\begin{algorithmic}[1]
\Procedure{GMC}{Expert $E$, max depth $d$}
\State target $t$, related attributes or entities $\mathbf{I} =$ \Call{Interface}{$E$}
\State Modes $\mathbf{M}=\emptyset$
\For{$i \in I$}
\State $\mathbf{Paths} =$ \Call{FindPaths}{$t, i, d$}
\For{$p \in \mathbf{Path}$}
\State $\mathbf{M}= \mathbf{M} \cup $ \Call{CreateMode}{$p$}
\EndFor
\EndFor
\State \Return $M$
\EndProcedure

\algrule

\Procedure{FindPaths}{target $t$, related attribute/entity $u$, find shortest path $isShortest$, max depth $d$}
\State $Solutions=\emptyset, Searched=\emptyset, ToExplore=\{t\}$
\While{$|ToExplore| > 0$ $\&\&$ $len(ToExplore.peek()) < d$}
\State $\mathbf{n} = \{x_1,r_1,x_2,r_2,...,r_{k-1},x_k\}$ $=ToExplore.dequeue()$
\For{$r \in \mathbf{R}_{x_k}$}
\For{Entity $y \neq x_k$ appearing in relation $r$}
\If{$\{\mathbf{n},r,y\} \in Searched$}
\State \textbf{continue}
\EndIf
\If{$y==u || u \in \mathbf{A}_y$}
\State $Solutions.append(\{\mathbf{n},r,y\})$
\If{$isShortest$}
\State \Return $Solutions$
\EndIf
\EndIf
\State $ToExplore.enqueue(\{\mathbf{n},r,y\})$
\EndFor
\EndFor
\EndWhile
\State \Return $Solutions$
\EndProcedure

\algrule

\Procedure{CreateMode}{Path $p$}
\State Modes $M=\emptyset$
\For{$\{x_i,r_i,x_{i+1}\}\in p$}
\For{Term $ t_j \in r_i$}
\If{$t_j==e_i$}
\State $t_j=+e_j$
\ElsIf{$t_j \in \mathbf{A}$}
\State $t_j=\#e_j$
\Else
\State $t_j=-e_j$
\EndIf
\EndFor
\State $M.append(r_i(t_0,t_1,...,t_n))$
\EndFor
\State \Return $M$
\EndProcedure
\end{algorithmic}
\end{algorithm}

\section{Human Guided Mode Construction}

Naive approaches for mode construction may allow for exhaustive search, enabling the ILP learner to find the best solution at the cost of a time intensive search process. Other approaches allow for one free variable for each atom considered. This restricts the search space, but ignores the fact that not all areas of the search space are equally important for a given target.

Alternatively, we consider guided construction of modes (\textsc{GMC}) for ILP where the human is assumed to be a domain expert and not an ILP expert. The expert is provided the structure of the domain in a graphical user interface that allows the expert to interact with the Entity-Relationship diagram. The target entity about which the model will be learned is marked and the expert is responsible for marking all of the attributes which are relevant to the target. Then, we find paths through entities and relations that are able to connect the target feature with all of the related entities\footnote{Note that ERDs represent entity sets/classes/types and not actual instances or entities to be precise. However, since in the context of our approach we never deal with instances, we use the term "entity" to denote entity classes for brevity} and their attributes. As we describe in more detail later, these paths are the basis for constructing the modes.

\subsection{An Illustrative Example}

\begin{figure*}
    \centering
    \subfigure[\textit{Target}: `Tenure'.   \textit{Informative/important}: `Grade', selected by user.]{
    \includegraphics[width=0.95\columnwidth]{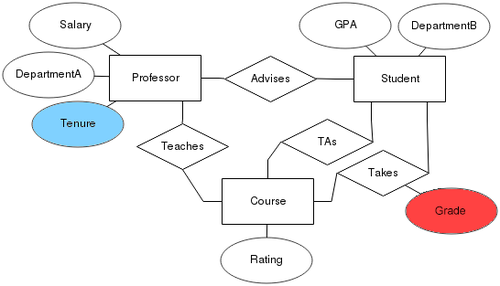}
    \label{fig:egERDSelected}
    }
    \subfigure[The equidistant shortest paths between Tenure and Grade.]{
    \includegraphics[width=0.95\columnwidth]{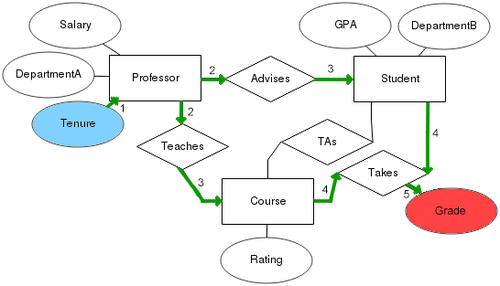}
    \label{fig:egERDDirected}
    }
    \caption{Illustrative example showing knowledge guided walks on the ERD, given in Figure~\ref{fig:egERD}, for mode construction.}
    \label{fig:my_label}
\end{figure*}

Consider a set of data involving professors, students, and courses, with some associated attributes and relationships between each. Figure~\ref{fig:egERDSelected} shows such an ERD where \textit{Grade} (marked in red) was identified by an expert as being an important attribute for predicting \textit{Tenure} (marked in blue).
\textsc{GMC} first connects the target concept to the related concepts by finding paths from one to another in the ER diagram. Figure~\ref{fig:egERDDirected} shows two paths that connect \textit{Tenure} to \textit{Grade}.

\begin{table*}
\begin{tabular}{l | c c c c c c}
     & Step 1 & Step 2 & Step 3 & Step 4 & Step 5 & Step 6 \\
    Path 1 & $Tenure$ & \hlightblue{Professor} & $Advises$ & \hlightred{Student} & $Takes$ & \hlightgreen{Grade} \\
    Modes 1 & $Tenure(+$\hlightblue{Prof}$)$ &  & $Advises(+$\hlightblue{Prof}$,-$\hlightred{Stud}$)$ & &  $Takes(+$\hlightred{Stud}$,-Course,\#$\hlightgreen{Grade}$)$ & \\
    Clause 1 & \multicolumn{6}{|c}{$Tenure(p)\wedge Advises(p,s)\wedge Takes(s,c,A+)$}
\end{tabular}
\caption{Each step of a path and corresponding modes generated by \textsc{GMC}.}
\label{tab:paths}
\end{table*}

Once these paths are established, variables can be set as being open (-), closed (+), or grounded (\#) based on the order in which entities (variables) appear. Since \textit{Tenure} is the target concept which everything should be learned with relation to, the conversion process begins with \texttt{Tenure(+Professor)}.

Modes are added to allow the ILP learner to search along the path. We show each step in one path and the corresponding modes that would be generated in Table~\ref{tab:paths}. Note that the entities and attributes are highlighted in different colors to show which arguments have the same type. The first time a type ($Student, Course$) is introduced along the path, the mode is set to $-$ allowing a free variable to be introduced during the search. Subsequently, appearances of a type have modes that are set to $+$, forcing a previous variable to be used during search. As $Grade$ is an attribute (as opposed to an entity), it will be grounded using the $\#$ mode.

Clause 1 in Table~\ref{tab:paths} gives an example of a clause that could be generated by an ILP system with the specified modes. The English interpretation of this rule is that tenure depends on the grades of students who are advised by a professor.

\subsection{The Algorithm}

The goal of \textsc{GMC} (Algorithm~\ref{algo:gmc}) is to guide the learner by constructing background knowledge based on input from a human user. This background knowledge consists of a set of modes that defines the search space for an ILP learner, enabling it to find a reasonable hypothesis efficiently. We have created a user interface that allows for a human domain expert to provide relevant attributes or entities for a given target (line \textbf{2}). \textsc{GMC} constructs modes that allow these relevant attributes or entities to appear in the model. Thus, the two key steps in \textsc{GMC} are 1) finding paths in the ER diagram (\textsc{FindPaths}) and 2) generating modes from those paths (\textsc{CreateMode}). We now discuss each of these steps.

\subsubsection{FindPaths}
Given the target $t$ and a relevant attribute or entity $u$, we find paths between them in the ER diagram. A path includes the set of entities and relationships which together relate $t$ to $u$. Each path $p=(t,r_t,x_1,r_1,x_2,r_2,...,r_{k-1},x_k)$ consists of attributes or entities ($\{x_i\}$) and relations ($\{r_j\}$). We explore the set of all paths in a breadth first manner starting from $t$. At each step, we select from among the shortest paths to expand. Assume $x_k$ is the current end of the selected path. We denote $\mathbf{R}_{x_k}$ as the set of relations in which entity $x_k$ appears. Path $p$ is then extended for each relation $r\in \mathbf{R}_{x_k}$ by creating a path for each entity that appears in $r$.

\textsc{GMC} finds a path when it reaches $u$ (if $u$ is an entity) or when it reaches an entity $y$ for which $u$ is an attribute ($u \in \mathbf{A}_y $). There are two potential settings corresponding to the number of paths to be found. If $isShortest$ is set to $true$, it will find a shortest path. Otherwise, it will find all paths up to a particular depth $d$. Our hypothesis is that finding all paths will yield background knowledge that encompass the best model while finding the shortest path will yield the most efficient set of modes that still allow the learner to find acceptable models. Note that the shortest path can be considered the most simple way to relate $t$ and $u$ and such simple knowledge is the basis for many learning algorithms.

\subsubsection{CreateModes}
Given a single path $p$ found by \textsc{FindPaths}, we now create a set of modes that will guide the search. As described previously, a mode is specified for a particular predicate. Each argument of the mode specifies the attribute type (defined by the structure of the ER diagram) as well as how new variables/constants can be introduced. For each relationship in the path $p$, we define a new mode. As mentioned earlier, the number and types of the arguments are defined by the structure of the ER diagram. We also assume that arguments corresponding to attribute values (e.g. the value of blood pressure or grade in a course) are considered as constants (\#). Thus, we only need to describe selecting between input/output variables.

For each pair of relations in the path connected through an entity ($(r_i,x_{i+1},r_{i+1})\in p$), we generate a mode $m_{r_{i+1}}$ for $r_{i+1}$. We denote $r^{x_{j}}_{k}$ to be the argument of relation $r_{k}$ that has associated type $x_{j}$. We set the argument $a=r^{x_{i+1}}_{i+1}$ as an input variable. All other arguments are set as an output variable ($\forall_{y\in Args(r_{i+1})\textbackslash a}$ $r^{y}_{i+1}$). Note that there could be multiple arguments with the same type ($|a|\geq 0$). If there are more than one, we generate a mode for each argument in $a$ as an input variable and treat all others as output variables.

The set of modes generated by \textsc{GMC} ($M$) can be used directly for ILP search. As \textsc{GMC} allows for the domain expert to only provide input on the ER diagram, no expertise in mode construction is required. We now describe our interface in more detail.

\subsection{The Interface}
\begin{figure}[ht]
    \centering
    \includegraphics[width=\columnwidth]{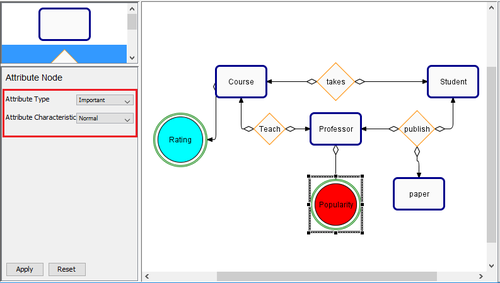}
    \caption{The Interface. It provides a drag-and-drop console with drop downs for annotating the ERD. As mentioned earlier: rectangles, diamonds, and ellipses/circles represent entities, relations, and attributes respectively. Here, ``Rating" is annotated as the \textit{target} and the ``Popularity" attribute is annotated as \textit{important}.}
    \label{fig:interface}
\end{figure}

The primary objective of our approach is to facilitate a domain expert, having limited understanding of ILP, in creating suitable modes as per the given problem. This necessitates an intuitive and user-friendly interface. We have developed a GUI (Figure~\ref{fig:interface}) that provides a user, having basic understanding of entities and relations, with the tools to build ER diagrams from scratch or load existing ones and annotate them with knowledge about targets and informative attributes/entities. The interface is designed for allowing the user to drag shapes and arrows to construct nodes and edges of a ER diagram as well as to select any node by double clicking on it to set its properties from the drop-down menus in the left pane. The properties include (1) whether the relation/attribute node is the \textit{\textbf{target}} (2) whether the attribute/relation is \textit{important/informative/predictive} and (3) if an attribute is multi-valued or binary.  Note that if the data is stored in a relational database, an ER diagram can be constructed automatically to some degree of fidelity. But, in most cases, the sanity or the quality of the ER-Diagrams are subject to the database designer's choice.

\section{Experiments}

\begin{figure*}[t]
\centering
\subfigure[CiteSeer Avg. Training Time]{
\centering
\includegraphics[width = 0.33\linewidth]{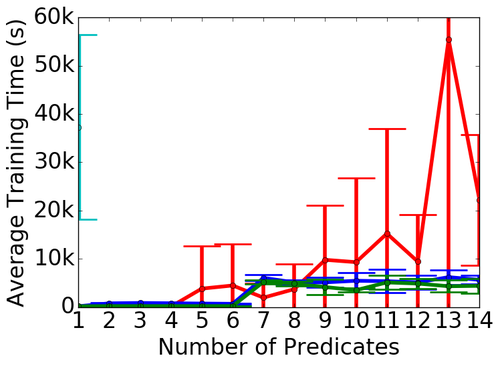}
\label{fig:citeseerTime}
}%
\subfigure[CiteSeer Avg. AUC ROC]{
\centering
\includegraphics[width = 0.33\linewidth]{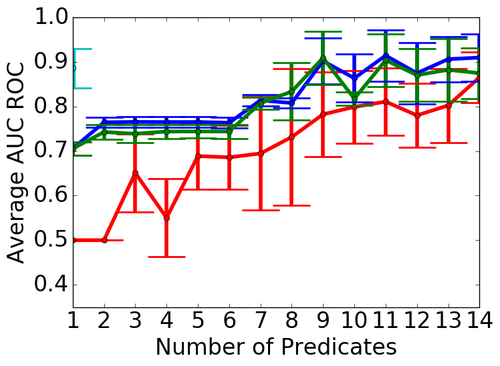}
\label{fig:citeseerROC}
}%
\subfigure[CiteSeer Avg. AUC PR]{
\centering
\includegraphics[width = 0.33\linewidth]{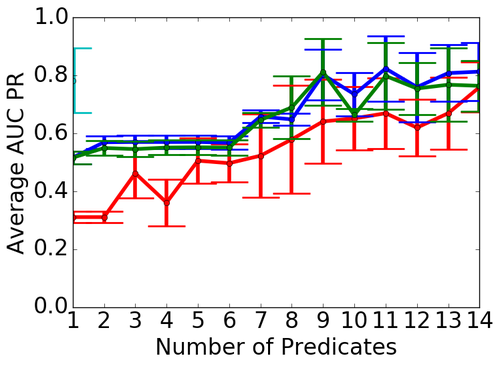}
\label{fig:citeseerPR}
}%
\\
\subfigure[WebKB Avg. Training Time]{
\centering
\includegraphics[width = 0.33\linewidth]{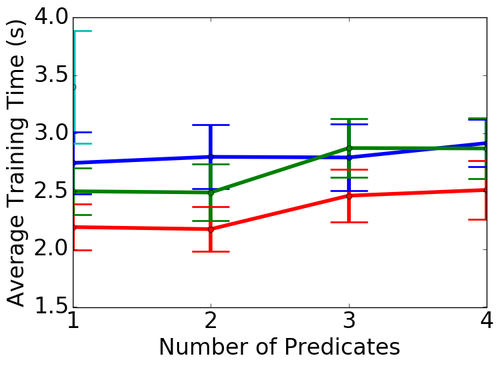}
\label{fig:webkbTime}
}%
\subfigure[WekKB Avg. AUC ROC]{
\centering
\includegraphics[width = 0.33\linewidth]{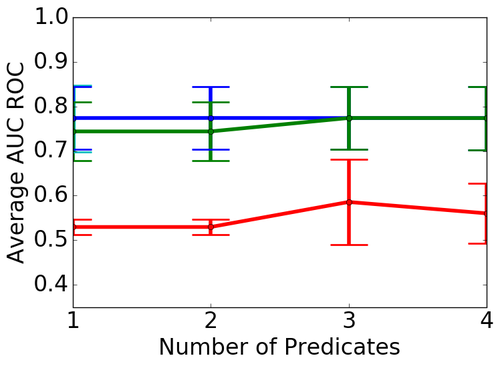}
\label{fig:webkbROC}
}%
\subfigure[WebKB Avg. AUC PR]{
\centering
\includegraphics[width = 0.33\linewidth]{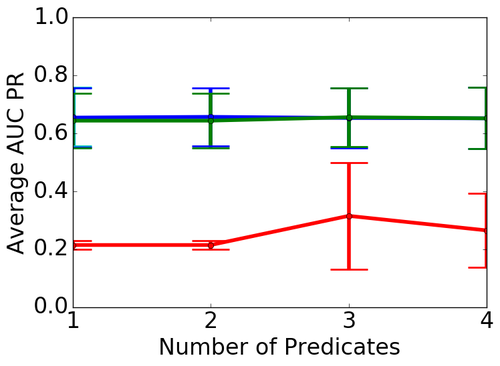}
\label{fig:webkbPR}
}%
\\
\subfigure{
\centering
\includegraphics[width=0.8\linewidth]{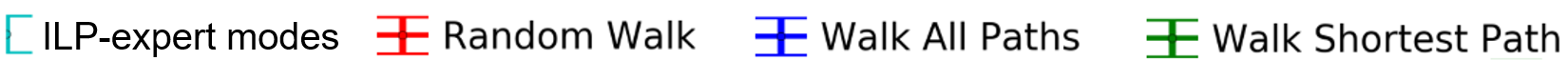}
}%
\caption{Results for CiteSeer and WebKB Datasets; Top row: Citeseer, Bottom row: WebKB. Left: Efficiency - Training time (lower is better), Middle \& Right: Performance - Average AUC ROC and AUC PR respectively (higher is better).}
\label{fig:results1}
\end{figure*}

\begin{figure*}[t]
\centering
\subfigure[Cora Avg. Training Time]{
\centering
\includegraphics[width = 0.33\linewidth]{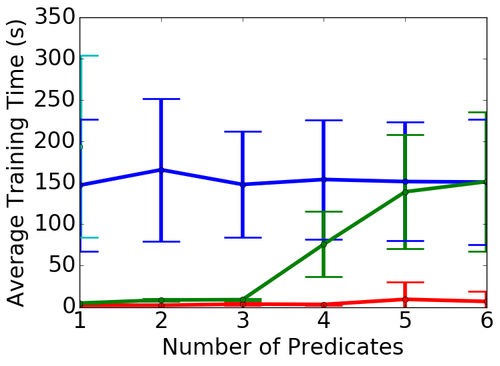}
\label{fig:coraTime}
}%
\subfigure[Cora Avg. AUC ROC]{
\centering
\includegraphics[width = 0.33\linewidth]{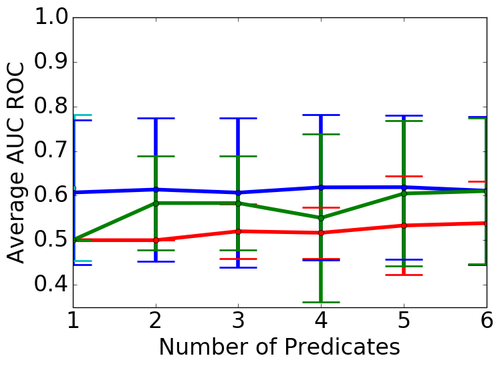}
\label{fig:coraROC}
}%
\subfigure[Cora Avg. AUC PR]{
\centering
\includegraphics[width = 0.33\linewidth]{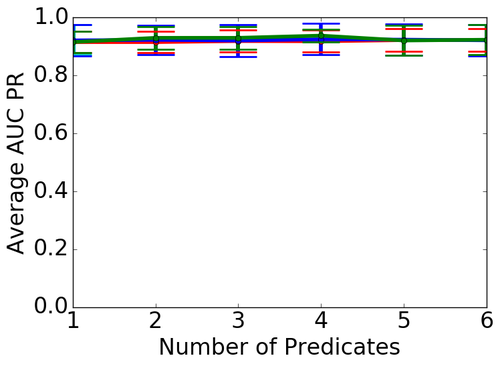}
\label{fig:coraPR}
}%
\\
\subfigure[IMDB Avg. Training Time]{
\centering
\includegraphics[width = 0.33\linewidth]{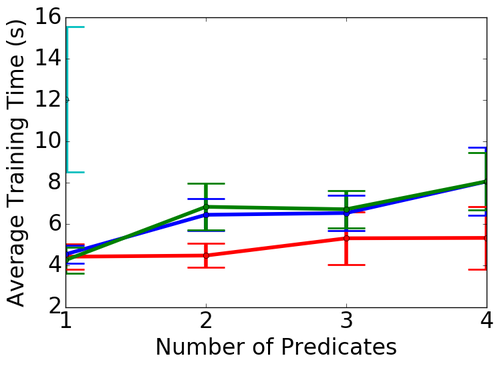}
\label{fig:imdbTime}
}%
\subfigure[IMDB Avg. AUC ROC]{
\centering
\includegraphics[width = 0.33\linewidth]{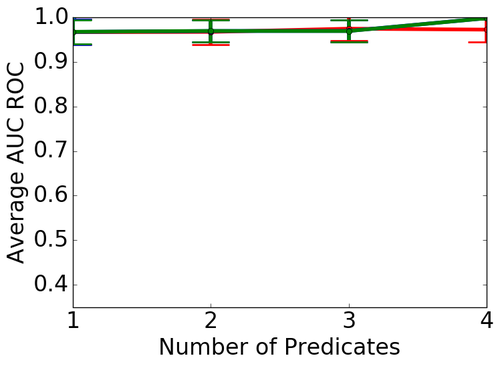}
\label{fig:imdbROC}
}%
\subfigure[IMDB Avg. AUC PR]{
\centering
\includegraphics[width = 0.33\linewidth]{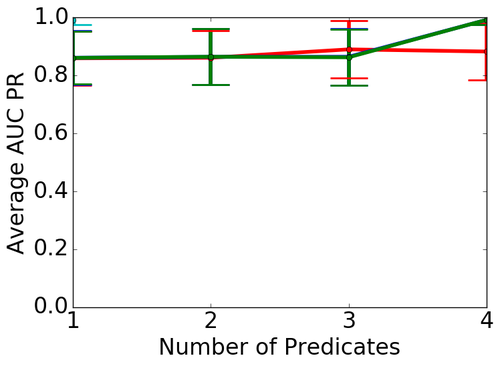}
\label{fig:imdbPR}
}%
\\
\subfigure{
\centering
\includegraphics[width=0.8\linewidth]{images/legend2.png}
}%
\caption{Results for Cora and IMDB Datasets; Top row: Cora, Bottom row: IMDB. Left: Efficiency - Training time (lower is better), Middle \& Right: Performance - Average AUC ROC and AUC PR respectively (higher is better).}
\label{fig:results2}
\end{figure*}

\begin{figure*}[t]
\subfigure[UW-CSE Avg. Training Time]{
\centering
\includegraphics[width = 0.33\textwidth]{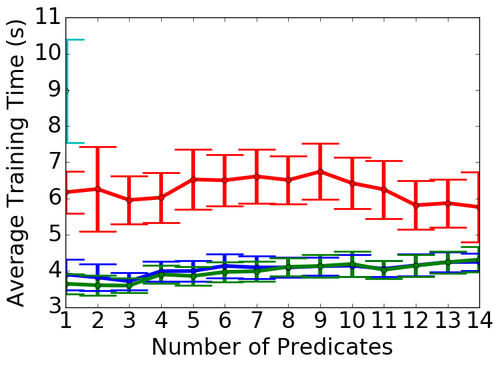}
\label{fig:uwcsetrainingtime}
}%
\subfigure[UW-CSE Avg. AUC ROC]{
\centering
\includegraphics[width = 0.33\textwidth]{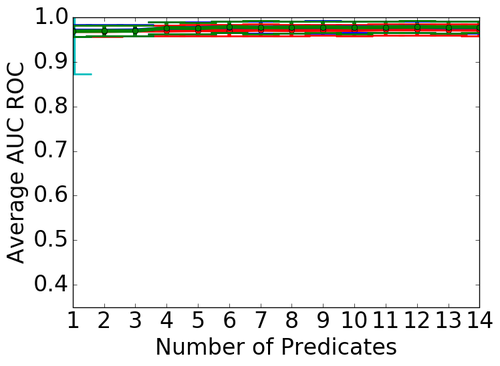}
\label{fig:uwcseaucroc}
}%
\subfigure[UW-CSE Avg. AUC PR]{
\centering
\includegraphics[width = 0.33\textwidth]{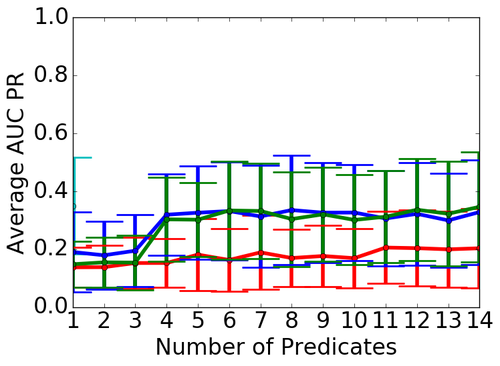}
\label{fig:uwcseaucpr}
}%
\\
\subfigure{
\centering
\includegraphics[width=0.8\linewidth]{images/legend2.png}
\label{fig:diagramlegend}
}%
\caption{UW-CSE results. Left: Efficiency - Training time (lower is better), Middle \& Right: Performance - Average AUC ROC and AUC PR respectively (higher is better).}
\label{fig:results3}
\end{figure*}

We pose the following questions to evaluate the effectiveness and efficiency of our approach (\textsc{GMC}) in background knowledge construction.
\begin{enumerate}
\item[\textbf{Q1}:] Does \textsc{GMC} facilitate the learner to optimally explore the hypothesis space (performance)?
\item[\textbf{Q2}:] Can \textsc{GMC} enhance efficiency via sufficiently constraining the search space?
\item[\textbf{Q3}:] Do simple (\texttt{Shortest Path}) modes generate robust models?
\item[\textbf{Q4}:] {What is the importance of human guidance?}
\end{enumerate}

We discuss our results based on two scenarios, (1) searching all paths from our target to our predicates, and (2) exploring the shortest paths. We compare our approach against two baselines: (a) modes encoded by an ILP expert\footnote{Discussion on manual mode construction is beyond the scope of this paper.},
and (b) mode construction based on depth-restricted random paths.

Note that \textit{a} achieves similar performance to walking from the target to every feature in the domain at a lower average training time. \textit{b} is inspired by the success of random walk algorithms that are capable of solving many challenging tasks~\cite{PRA}.

The system has two components, (a) a platform-independent GUI component for creation and annotation of ER diagrams and (b) the mode construction from the annotated diagram which is designed to be compatible with any ER diagramming tool given a common intermediate representation.

We have used the state-of-the-art ILP structure/parameter learning framework Relational Functional Gradient Boosting~\cite{natarajan10} as the test-bed for evaluating the quality of automatically constructed modes.

\subsection{Domains}

We use four standard ILP/PLM datasets, namely CiteSeer, WebKB, Cora, and IMDB, for an empirical evaluation of our automatic mode construction system. \textit{``facts"} refers to the evidence (all the relations between different objects that are true in the given domain) and \textit{``examples"} refers the total number of positive and negative (true and false) target relations/attributes/features across each cross-validation fold.

\noindent \textbf{CiteSeer}~\cite{poon07} dataset was created for information extraction and citation matching. It has 121,891 facts and 116,679 examples split across four cross-validation folds, each corresponding to a different topic. Our goal was to predict which field the title of the paper corresponded to (\texttt{infield\_ftitle}), and the fourteen other predicates are based on tokens and their relative positions in a document.

\noindent \textbf{WebKB}~\cite{bottomupmln07} is a consolidated dataset of links among departmental web pages from four universities (Cornell University, University of Texas, University of Washington, and University of Wisconsin) each grouped into one of four cross-validation folds. It has 1912 facts and 747 examples, where the target is to predict \texttt{faculty} based on several predicates (\texttt{courseProf}, \texttt{courseTA}, \texttt{project}, and \texttt{samePerson}).

\noindent \textbf{Cora}~\cite{poon07}, like CiteSeer, is about citation matching, with the key difference being the type of relations that are captured. The dataset consists of 6,541 facts and 62,715 examples split into five cross-validation folds. The target is to predict if 2 citations have the same author (\texttt{sameAuthor}).

\noindent \textbf{IMDB}~\cite{bottomupmln07} represents relations between movies and the people who work on them, as well as several attributes of such movies and people. People can either be an actor or a director (mutually exclusive), and the goal is to predict whether an actor worked under a certain director (\texttt{workedUnder}). In total there are 664 facts and 5794 examples.

\noindent \textbf{UW-CSE} is an anonymized representation of the staff and students of five computer science departments distributed across five cross-validation folds; consisting of 5121 facts and 94,000 examples. The goal is to predict who advises whom (\texttt{advisedby}).

\subsection{Experimental Setup}

Our \textsc{GMC} algorithm has two settings, \texttt{Walk All Paths} for walking all paths on the graph from the user-specified target to each selected feature, and \texttt{Walk Shortest Path} for finding only a shortest path from the target to each selected feature.

Experiments were performed on a server with twenty Intel Xeon E5-2690 CPUs clocked at 3.00GHz with no other processes on the server which might interfere with training time. To compare performance for each method, we report the mean and standard deviation of the \textit{training time}, \textit{AUC ROC}, and \textit{AUC PR} across 5 cross-validation folds and 10 independent runs for every dataset and number of features. The settings (namely: negative:positive ratio and \#literals at each tree-node) of the underlying PLM learner, `RFGB', were kept consistent across all the evaluated approaches and 10 trees were learned in each case.

Features (attributes/relations) the expert annotates as important/informative are arranged in the order in which they were selected. In the experimental results (Figures~\ref{fig:results1},~\ref{fig:results2} \& \ref{fig:results3}), the \textbf{x-axis} represents this ordering, and the respective values for each point represents the performance of a horizontal slice of all predicates up to and including that point. This shows how each additional predicate influences performance/training time.

\subsection{Experimental Results}

From Figures~\ref{fig:results1}, \ref{fig:results2} and ~\ref{fig:results3} we observe that both settings of our \textsc{GMC} algorithm outperforms \textit{Random Walk} in 3 of the datasets (CiteSeer/WebKB in AUC ROC and CiteSeer/WebKB/UW-CSE in AUC PR). The difference is more pronounced earlier in the learning curve when fewer paths are being found.  As expected, when the number of paths increase, the performance of \textit{Random Walk} often approaches \textsc{GMC}. Both of our \textsc{GMC} approaches are capable of matching the performance of \textit{ILP-expert modes}, often with very few informative predicates marked (except in the case of CiteSeer which requires additional marked predicates). Thus, our \textsc{GMC} methods generate modes that facilitate effective ILP search (\textbf{Q1}).

While our \textsc{GMC} approaches generate high performance, they also constrain the search space to allow for efficient models to be learned. The training time of \textit{Random Walk} varies: it is lower than our approaches in three datasets (WebKB/Cora/IMDB) and higher in CiteSeer and UW-CSE. Even though \textit{Random Walk} is more efficient, it is less effective (WebKB/Cora) than our approaches. When compared to the \textit{ILP-expert modes}, our \textsc{GMC} approaches are significantly more efficient in all domains except Cora, where \textit{Walk All Paths} performs similarly to \textit{ILP-expert modes}. Overall, both of our \textsc{GMC} approaches are capable of learning more efficient models than the baseline while also achieving high performance (\textbf{Q2}).

While both variants of our \textsc{GMC} algorithm (\textit{Walk All Paths} and \textit{Walk Shortest Path}) compare favorably to the other baselines, we now discuss their differences. Intuitively, \textit{Walk Shortest Path} should have an efficiency advantage over \textit{Walk All Paths}. This is demonstrated in two domains (WebKB/Cora) where \textit{Walk Shortest Path} achieves similar performance to \textit{Walk All Paths} while having significantly lower training time. In all other domains, both variants perform similarly. This suggests that the shortest explanation is often sufficient and allows for a robust and efficient search (\textbf{Q3}).

To better comprehend the role of human guidance (\textbf{Q4}), let us consider two---not necessarily distinct---aspects. Primarily, human guidance acts as search space constraints for the ILP learner to efficiently search for models. Hence, careful encoding of modes is necessary to achieve comparable, at times better, performance than a super-exponential exhaustive search. \textit{Random Walks} can manage to reduce the search space by working with randomly sampled regions. However, as the results illustrate (Figures~\ref{fig:webkbROC},~\ref{fig:webkbPR}, etc.), it may not result in robust models. The other aspect is knowledge about what the most important features/nodes are in automatic mode construction. The empirical results illustrate that, across all datasets and all empirical measurements, there exists a convergence point where including additional guidance (annotations of important features) no longer leads to better performance. \textit{IMDB} requires all four predicates to be taken into account; but in \textit{CiteSeer}, \textit{WebKB}, and \textit{Cora}: performance no longer improves after predicates \textbf{9}, \textbf{3}, and \textbf{1}, respectively. In all cases except \textit{Cora}, training time continues to increase slightly while overall performance stabilizes. The domain expert is essential for providing the initial ER model as well as annotating what the most important features/nodes are.

\section{Conclusion}

We considered the problem of capturing domain expert knowledge in the context of learning first-order probabilistic models. We developed a solution based on entity relationship diagrams that allows the domain expert to provide relevant knowledge effectively for making the search process efficient. Our solution is inspired by the observation that most  probabilistic logic models can be seen as learning a probabilistic model over a relational graph in the lines of probabilistic relational models~\cite{prm}
and probabilistic entity-relational models~\cite{daper}. Given this observation, the domain expert identifies relevant nodes in the ER diagram which translates to providing appropriate modes for a clause learning system. Our experiments on standard PLM domains demonstrate the effectiveness of our proposed approach.
Extending this system to actively solicit advice as needed~\cite{odom2016ECML} is a possible future direction. Allowing for incomplete/noisy and even competing advice is another direction. Finally, extending the interface to allow for knowledge capture in other learning frameworks such as sequential decision-making in relational models, relational deep networks, and other relational models remain an interesting direction for future research.

\section*{Acknowledgements}
Mayukh Das and Sriraam Natarajan gratefully acknowledge the support of the CwC Program Contract W911NF-15-1-0461 with the US Defense Advanced Research Projects Agency (DARPA) and the Army Research Office (ARO).
Phillip Odom and Sriraam Natarajan acknowledge the support of the Army Research Office (ARO) grant number W911NF-13-1-0432 under the Young Investigator Program.

\bibliographystyle{abbrv}
\bibliography{modeconstruction,biblio}
\end{document}